# Stock Price Prediction using Multi-Faceted Information based on Deep Recurrent Neural Networks

Lida Shahbandari
Department of Computer Engineering, North Tehran Branch, Islamic Azad University, Tehran, Iran
lida.shahbandari@gmail.com

Elahe Moradi
Department of Electrical Engineering, Yadegar-e-Imam Khomeini (RAH) Shahre Rey Branch, Islamic Azad University, Tehran, Iran
elahemoradi@iau.ac.ir

Mohammad Manthouri
Department of Electrical and Electronic Engineering
Shahed University
Tehran, Iran
mmanthouri@shahed.ac.ir

*Abstract*— Accurate prediction of stock market trends is crucial for informed investment decisions and effective portfolio management, ultimately leading to enhanced wealth creation and risk mitigation. This study proposes a novel approach for predicting stock prices in the stock market by integrating Convolutional Neural Networks (CNN) and Long Short-Term Memory (LSTM) networks, using sentiment analysis of social network data and candlestick data (price). The proposed methodology consists of two primary components: sentiment analysis of social network and candlestick data. By amalgamating candlestick data with insights gleaned from Twitter, this approach facilitates a more detailed and accurate examination of market trends and patterns, ultimately leading to more effective stock price predictions. Additionally, a Random Forest algorithm is used to classify tweets as either positive or negative, allowing for a more subtle and informed assessment of market sentiment. This study uses CNN and LSTM networks to predict stock prices. The CNN extracts short-term features, while the LSTM models long-term dependencies. The integration of both networks enables a more comprehensive analysis of market trends and patterns, leading to more accurate stock price predictions.

*Keywords— Stock Price Prediction, Deep Learning, Sentiment Analysis, Long Short-Term Memory, Convolutional Neural Network*

## I. INTRODUCTION

The stock market is a complex and dynamic system, characterized by volatility, uncertainty, and non-linear relationships between various economic and financial factors. Its behavior is influenced by a multitude of factors, including macroeconomic indicators, company-specific news, and investor sentiment, making it challenging to predict and analyze. Accurate prediction of stock prices is crucial for investors, analysts, and portfolio managers, as it enables them to make informed investment decisions, minimize risk, and maximize returns [1]. Effective prediction models can also provide a competitive edge in the financial market, allowing individuals and organizations to stay ahead of the curve and capitalize on emerging trends and opportunities. Traditional approaches to stock price prediction include fundamental analysis, which examines a company's financial statements, management team, and industry trends to estimate its intrinsic value [2]. Technical analysis is another traditional approach, which involves identifying patterns and trends in historical price data to forecast future price movements. Statistical models, such as autoregressive integrated moving average (ARIMA) and generalized autoregressive conditional heteroskedasticity (GARCH), are also used to predict stock prices based on historical data [3]. [4] examines the impact of Bitcoin Exchange-Traded Funds (ETFs) on traditional financial markets, focusing on liquidity, volatility, and investor behavior, to understand the effects of this emerging financial product on established markets. This revolutionary robot [5], designed to mimic human size and proportions, showcases superior capabilities, unwavering stability, and sophisticated equilibrium management, rendering it a premier tool for innovators and researchers in the humanoid robotics sector.

Machine learning approaches, such as decision trees, random forests, and support vector machines, can be used to predict stock prices by identifying complex patterns and relationships in large datasets. Deep learning models, including recurrent neural networks (RNNs) and convolutional neural networks (CNNs), can learn to extract features and make predictions based on sequential and spatial data, such as time series data and financial news articles. Additionally, hybrid models that combine machine learning and deep learning techniques, such as LSTM-CNN models, can be used to predict stock prices by leveraging the strengths of both approaches [6]. Long Short-Term Memory (LSTM) networks, a type of RNN, have emerged as a potent tool for time series forecasting, particularly in stock price prediction. By leveraging their ability to learn long-term dependencies, LSTMs can uncover hidden patterns and trends in historical stock prices and trading volumes, enabling them to make accurate predictions of future price movements. For instance, an LSTM network can identify the correlation between specific news events and subsequent stock price increases, or recognize the predictive power of certain technical indicators [7]. By harnessing these insights, LSTMs can generate reliable forecasts of future stock prices, making them particularly well-suited for long-term predictions. CNNs, traditionally used for image and signal processing, have also proven effective in Natural Language Processing (NLP) tasks, such as analyzing financial news articles and social media posts. In the context of stock price prediction, CNNs can extract valuable features from unstructured data, including sentiment and event-based features. For example, a CNN can be trained to analyze financial news articles, extracting features like sentiment scores, entity recognition, and topic modeling. These features can then be fed into a machine learning model, such as a random forest or support vector machine, to predict stock prices. Additionally, CNNs can analyze social media posts, extracting features like sentiment, volume, and topic modeling, which can also inform stock price predictions. CNNs are particularly effective in short-term prediction, where they can quickly identify patterns and trends in recent data, making them ideal for predicting stock price movements over shorter time horizons. By tapping into the strengths of CNNs in NLP, researchers can uncover valuable insights from unstructured data, leading to more accurate stock price predictions[8].

NLP has become a crucial tool in stock market prediction, particularly in the realm of sentimental analysis. By leveraging NLP techniques, such as text classification, topic modeling, and named entity recognition, researchers can extract valuable insights from unstructured data, including financial news articles, social media posts, and company reports, to quantify the sentiment of market participants [9]. Sentiment analysis is a subfield of NLP that focuses on identifying and extracting subjective information from text data. Sentimental analysis can be used to analyze earnings call transcripts, social media posts, and financial news articles to extract sentiment scores that can be used to predict stock prices, providing a deeper understanding of market sentiment and enabling more accurate predictions about future stock price movements. Furthermore, this approach can also help investors identify potential investment opportunities and mitigate risks by detecting early signs of market shifts and trends [10].

This research presents a novel methodology for predicting stock prices in the stock market, which involves the integration of CNN and LSTM networks, in conjunction with sentiment analysis of social network data, to facilitate more accurate and informed investment decisions. Here are the contributions of this paper:

• **Fusion of Candlestick and News Data:** This methodology integrates candlestick data and tweet-derived information to facilitate a more precise and exhaustive examination of market trends and patterns in stock price prediction. The amalgamation of these disparate data sources enables the capture of both technical and fundamental factors that exert influence on stock prices, thereby providing a more comprehensive understanding of market dynamics. Furthermore, a Random Forest algorithm is employed to categorize the sentiment of each tweet as either negative or positive, thereby facilitating a more nuanced analysis of market sentiment.

• **Novel Integration of CNN and LSTM:** This study proposes a novel approach that integrates CNN and LSTM networks to predict stock prices, enabling the extraction of more accurate patterns and better understanding of market trends. The CNN component is used to extract spatial hierarchies of features from data, while the LSTM component is used to model temporal dependencies in the data. The integration of these two networks allows for a more comprehensive analysis of market trends and patterns, leading to more accurate stock price predictions.

The paper is structured as follows: Section II provides an overview of the stock market forecasting dataset, while Section III delves into the methodology employed, which integrates NLP-based sentiment analysis with a CNN-LSTM architecture. The results of the simulation, including the training and evaluation of the hybrid approach, are presented in the next section. The paper concludes with a summary of the main findings and contributions.

## II. STOCK MARKET FORECASTING

This section showcases the Stock Market Database as a useful dataset for deciphering market trends and predicting future movements. By leveraging this extensive resource, researchers, traders, and financial analysts can gain valuable insights into the application of deep learning techniques in market analysis. We examine the latest research on cutting-edge methods such as NLP, CNNs, and LSTMs, and their potential to enhance stock market forecasting.

### A. Stock Market Dataset

The daily time frame offers a distinct advantage in the stock market, providing a broader perspective on market trends and dynamics. Unlike shorter time frames, which can be influenced by fleeting market sentiment and noise, the daily time frame allows traders and analysts to capture more significant and sustained trends, making it an ideal choice for swing trading and position trading strategies. By focusing on the daily time frame, traders can filter out intraday volatility and noise, and instead, identify more reliable and robust trading opportunities, ultimately leading to more informed and effective decision-making. Additionally, the daily time frame is less susceptible to overnight risks and news-driven events, making it a more stable and predictable choice for traders seeking to maximize returns in a fast-paced and volatile market environment.

In this study, we conducted sentiment analysis on Twitter data related to Amazon and Tesla, two of the most influential and widely followed companies in the world. Our goal was to understand the public's perception and sentiment towards these companies, which can have a significant impact on their stock prices. We collected a large dataset of tweets related to Amazon and Tesla, and applied various NLP techniques to analyze the sentiment of the tweets. Our results showed that the sentiment of tweets towards Amazon was generally positive, with a significant proportion of tweets expressing admiration for the company's innovative products and services. On the other hand, the sentiment of tweets towards Tesla was more mixed, with some tweets expressing enthusiasm for the company's electric cars and sustainable energy solutions, while others expressed concerns about the company's financials and production challenges. In addition to sentiment analysis, we also applied deep learning approaches to predict the candlestick patterns of Amazon and Tesla stocks. Candlestick patterns are a popular technical analysis tool used to predict stock price movements, and deep learning models have been shown to be effective in predicting these patterns.

## B. Literature Survey

RNNs, particularly LSTM networks and 1-dimensional CNNs, have proven to be highly effective in stock market prediction due to their unique ability to process sequential data and capture complex temporal dependencies. By leveraging these architectures, researchers and practitioners can better model the dynamic and non-linear relationships inherent in financial time series data, ultimately leading to more accurate and reliable predictions. [11] proposes a multi-parameter forecasting approach for stock time series data using LSTM and deep learning models, demonstrating improved forecasting accuracy and robustness by incorporating multiple parameters and leveraging the strengths of deep learning techniques. [12] presents a novel graph-based CNN-LSTM stock price prediction algorithm that incorporates leading indicators, demonstrating improved forecasting accuracy and robustness by leveraging the strengths of graph convolutional networks and RNNs to capture complex temporal and spatial relationships in stock market data. [13] proposes a CNN-LSTM stock prediction model that utilizes a genetic algorithm for optimization, demonstrating improved forecasting accuracy and robustness by leveraging the strengths of deep learning and evolutionary optimization techniques to select optimal model parameters and hyperparameters. [14] This paper presents a novel CNN-BiLSTM-Attention model for stock price prediction, demonstrating improved forecasting accuracy by leveraging the strengths of CNNs, bidirectional LSTM networks, and attention mechanisms to capture complex patterns and relationships in stock market data. [15] proposes a hybrid CNN-LSTM model for portfolio performance optimization, demonstrating its effectiveness in stock selection and portfolio optimization through a case study, and showcasing the potential of deep learning techniques to improve investment decision-making and portfolio returns. [16] presents a novel hybrid deep learning model that combines stacked bidirectional LSTM and evolutionary-optimized CNN, demonstrating improved stock price prediction accuracy and robustness by leveraging the strengths of both RNNs and CNNs.

Sentiment analysis, a key application of NLP, involves the use of machine learning algorithms to automatically determine the emotional tone or sentiment conveyed by text data, such as social media posts, customer reviews, or financial news articles [17]. By leveraging NLP techniques, including tokenization, part-of-speech tagging, and named entity recognition, sentiment analysis can effectively identify and classify text as positive, negative, or neutral, providing valuable insights into public opinion, market trends, and customer behavior. [18] proposes a novel approach to stock market sentiment analysis using BERT, a pre-trained language model, and demonstrates its effectiveness in capturing nuanced sentiment expressions and improving prediction accuracy. [19] presents a hybrid approach that combines data mining techniques with news sentiment analysis to predict stock market behavior, demonstrating the potential of integrating multiple data sources and analytical methods to improve prediction accuracy. [20] proposes a novel learning-based approach that combines technical indicators and social media sentiment analysis to predict stock trending, demonstrating improved prediction accuracy and highlighting the potential of integrating multiple data sources for informed investment decisions. [10] employs machine learning sentiment analysis to investigate the impact of COVID-19 news on stock market reactions, revealing significant correlations between news sentiment and market volatility, and providing insights into the emotional and psychological factors driving investor behavior during the pandemic.

## III. METHODOLOGY

The third section of the paper is divided into three key components: NLP, sentiment analysis using a random forest classifier, and a proposed deep learning approach for predicting stock market prices.

### A. Natural Language processing (NLP)

TF-IDF is a widely used statistical technique in NLP for weighting the importance of words in a document based on their frequency of occurrence. The technique is designed to capture the relevance of a word in a document by considering both its local and global importance. The Term Frequency (TF) component of TF-IDF measures the frequency of a word in a document. It is calculated as the ratio of the number of times a word appears in a document to the total number of words in the document. TF is a measure of the local importance of a word in a document. The Inverse Document Frequency (IDF) component of TF-IDF measures the rarity of a word across a collection of documents. It is calculated as the logarithm of the total number of documents in the collection divided by the number of documents containing the word. IDF is a measure of the global importance of a word in a collection of documents.

Let D be a collection of documents, and let w be a word in the vocabulary. The TF-IDF score for word w in document d is calculated as:

$$TF - IDF(w, d) = (tf(w, d) \times idf(w)) \quad (1)$$

where $TF$ is the Term Frequency and $IDF$ is the Inverse Document Frequency and the corresponding for are shown as follows:

- $tf(w, d) = (number\ of\ occurrences\ of\ w\ in\ d) / (total\ number\ of\ words\ in\ d)$
- $idf(w) = \log(N / (number\ of\ documents\ containing\ w)$

$N$ is the total number of documents in the collection.

TF-IDF possesses several desirable properties that make it a popular choice for NLP tasks, including term weighting, document normalization, robustness to noise, and scalability. These properties enable TF-IDF to effectively capture the importance of words in a document, reduce the impact of document length and word frequency, and down-weight irrelevant words, ultimately providing a robust and efficient feature extraction technique for various NLP applications.

After preprocessing the tweets using TF-IDF, the resulting feature matrix is fed into a Random Forest classifier to classify the tweets into their respective categories. The Random Forest algorithm, known for its robustness and accuracy, is well-suited for handling the high-dimensional feature space generated by TF-IDF. By combining the strengths of TF-IDF and Random Forest, we aim to achieve a highly accurate classification model that can effectively distinguish between different types of tweets. The Random Forest classifier is trained on the TF-IDF feature matrix, and its performance is evaluated using metrics such as accuracy, precision, recall, and F1-score.

### B. Random Forest Classifier

A Random Forest is an ensemble of decision trees. The random forest formula is based on the bagging (Bootstrap Aggregating) algorithm, which combines the predictions of multiple decision trees. The decision tree formula is:

$$T(x) = argmax(g(x, \theta)) \qquad (2)$$

where $g(x, \theta)$ is the decision function, $\theta$ is the parameter vector. The decision function $g(x, \theta)$ is defined as:

$$g(x, \theta) = \sum (w_j \cdot \phi_j(x)) \qquad (3)$$

where $w_j$ is the weight of the $j-th$ feature, $\phi_j(x)$ is the feature function. In the context of decision trees, the decision function $g(x, \theta)$ typically represents the process of making decisions based on the features of the input $x$, guided by the parameters $\theta$ of the tree (such as the split points and feature thresholds). The $argmax$ operation in the decision tree formula then selects the output with the highest decision function value.

Let's denote the random forest as $RF = \{T_1, T_2, \ldots, T_m\}$, where $T_i$ is the $i-th$ decision tree. The Random Forest formula is:

$$RF(x) = \frac{1}{m} \sum T_i(x) \qquad (4)$$

where $RF(x)$ is the is the prediction of the random forest model for the input $x$. $m$ is the total number of decision trees in the forest. The Gini impurity is a measure of the purity of a node in a decision tree. The Gini impurity formula is:

$$Gini(D) = 1 - \sum p_i^2 \qquad (5)$$

where $Gini(D)$ is the Gini impurity, $p_i$ is the proportion of instances in the $i-th$ class, $D$ is the data,

$$Entropy(D) = -\sum (p_i \cdot log_2(p_i)) \qquad (6)$$

where $Entropy(D)$ is the entropy, $p_i$ is the proportion of instances in the $i-th$ class. These measures are used to evaluate the quality of a split in the decision tree or random forest. A lower value of Gini impurity or entropy means that a node is purer, with a majority of instances belonging to the same class, while a higher value means that the instances in the node are mixed up with different classes.

This method utilizes a Random Forest classifier to determine the sentiment of tweets and assess their impact on a given index. The approach involves the following steps:

- Tweet Sentiment Classification: A Random Forest classifier is trained on a dataset of labeled tweets, where each tweet is assigned a sentiment score of +1 (positive) or -1 (negative). The classifier learns to predict the sentiment of new, unseen tweets based on their features.
- Tweet Sentiment Scoring: The trained Random Forest classifier is applied to a new set of tweets, and each tweet is assigned a sentiment score of +1 (positive) or -1 (negative).
- Index Impact Calculation: The sentiment scores of the tweets are aggregated to calculate the overall impact of the tweets on the index.

The calculation is performed as follows:

$$I = \sum S_t \qquad (7)$$

where $S_t$ is the sentiment score of the $t-th$ tweet. Let $I$ be the set of tweets, and S be the sentiment score of each tweet (S ∈ {+1, -1}).

### C. Proposed Deep Learning Model

The stock market is a complex and dynamic system, making it challenging to predict its behavior. However, with the advent of deep learning techniques, particularly LSTM networks, it has become possible to develop accurate predictive models for stock market forecasting. By leveraging the sequential nature of financial time series data, LSTM networks can learn to identify patterns and relationships between historical stock prices, trading volumes, and other relevant factors. Our proposed model utilizes an LSTM architecture to predict stock market trends, enabling investors and traders to make informed decisions based on data-driven insights. By analyzing large datasets of historical stock market data, the LSTM model can learn to recognize patterns and anomalies, providing a robust framework for predicting future stock prices and identifying potential investment opportunities. The LSTM layer, a key component of the model, can be mathematically represented by a set of equations.

$$f_t = \sigma(W_f \cdot [h_{t-1}, x_t] + b_f) \qquad (8)$$
$$i_t = \sigma(W_i \cdot [h_{t-1}, x_t] + b_i) \qquad (9)$$
$$\tilde{C}_t = tanh(W_C \cdot [h_{t-1}, x_t] + b_C) \qquad (10)$$
$$C_t = f_t \times C_{t-1} + i_t * \tilde{C}_t \qquad (11)$$
$$o_t = \sigma(W_o \cdot [h_{t-1}, x_t] + b_o) \qquad (12)$$
$$h_t = o_t \times tanh(C_t) \qquad (13)$$

The LSTM layer is characterized by a triplet of gating mechanisms, comprising the forget gate ($f_t$), input gate ($i_t$), and output gate ($o_t$), which modulate the information flow between the cell state ($C_t$) and hidden state ($h_t$). At each time step $t$, the input $x_t$ is processed, and the sigmoid activation function (σ) is employed to introduce non-linear transformations within the gates. The computation of the gates and states is facilitated by learnable parameters, including weight matrices (W) and bias vectors (b).

CNN1D is a powerful tool for predicting stock prices by analyzing historical data. This architecture is designed to capture local patterns and trends in the data, making it well-suited for time series forecasting tasks. By applying a set of filters to the input data, the CNN1D model extracts relevant features and learns to recognize patterns that are indicative of future price movements. The model's ability to handle large datasets and robustness to noise make it an effective tool for stock market prediction, allowing investors and traders to make informed decisions based on data-driven insights.

$$X_{conv_t} = \sigma(W * X[t-k:t+k] + b) \qquad (14)$$

Where $X[t]$ is the input time series data at time step $t$, $k$ is the kernel size (i.e. the number of time steps used in the convolution).

The outputs of the LSTM layer and the CNN1D layer are combined into a single vector, $Y_t$, by concatenating the two outputs together.

$$Y_t = W_{CNN} \times X_{conv_t} + W_{LSTM} \times h_t + b \quad (15)$$

where:
- $Y_t$ is the output at time step $t$
- $\sigma$ is the activation function (e.g. ReLU or tanh)
- $W_{CNN}$ is the weight matrix for the CNN output
- $X_{conv_t}$ is the output of the CNN at time step $t$
- $h_t$ is the hidden state of the LSTM at time step $t$
- $b$ is the bias term

The proposed method uses a combination of candlestick data Open, High, Low, Close (OHLC) and sentiment analysis data to predict stock prices. The input features of the network are:
1. **Open**: The opening price of the stock.
2. **High**: The highest price of the stock.
3. **Low**: The lowest price of the stock.
4. **Close**: The closing price of the stock.
5. **Sentiment**: The sentiment analysis result obtained using a Random Forest classifier, which represents the emotional tone of the text data related to the stock.

The goal of the network is to predict the future stock prices based on the historical OHLC data and sentiment analysis results. By combining these two types of data, the network aims to capture both the technical and fundamental aspects of the stock market, and make more accurate predictions.

## IV. SIMULATION RESULTS

In this section, we evaluate the performance of the proposed model using historical OHLC data and sentiment analysis to predict the stock prices of two of the most influential and widely followed companies in the world: Amazon (AMZN) and Tesla (TSLA). The predictions of AMZN and TSLA using proposed method are shown in Figures 1-2, and the performance indices are shown in Table 1-2. Additionally, the results of sentimental analysis using the Random Forest analysis are shown in Figure 3-4.

R-squared, MSE, RMSE, and MSLE are four commonly used metrics to evaluate the performance of a regression model. R-squared measures the goodness of fit of the model, with values ranging from 0 to 1. MSE and RMSE measure the average magnitude of errors made by the model, with lower values indicating better fit. MSE calculates the average squared difference between predicted and actual values, while RMSE takes the square root of MSE. MSLE, on the other hand, measures the average squared difference between the logarithms of predicted and actual values, focusing on relative errors. Together, these metrics provide a comprehensive understanding of a model's performance, with R-squared assessing fit, MSE and RMSE evaluating error magnitude, and MSLE considering relative errors.

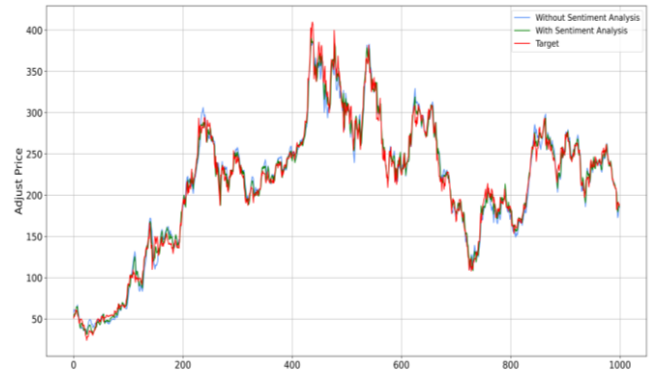

Figure 1: Predicted AMZN price using the with and without the proposed Sentiment Analysis.

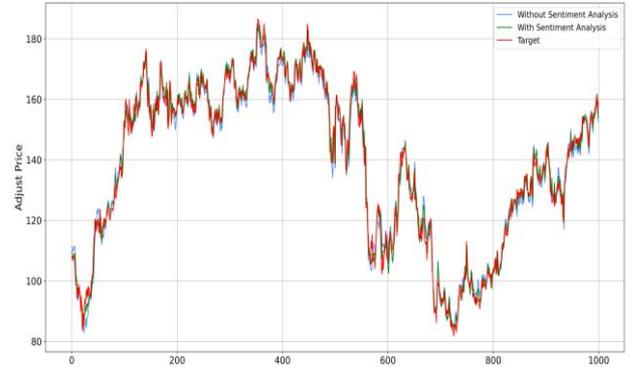

Figure 2: Predicted TSLA price using the with and without the proposed Sentiment Analysis.

Table 1. Performance AMZN price using the with and without the proposed Sentiment Analysis.

|  | MSE | RMSE | R − Square | MSLE |
| --- | --- | --- | --- | --- |
| Without NLP | 12.53 | 3.540 | 0.987 | 0.0008 |
| With NLP | 8.92 | 2.987 | 0.991 | 0.0005 |

Table 2. Performance TSLA price using the with and without the proposed Sentiment Analysis.

|  | MSE | RMSE | R − Square | MSLE |
| --- | --- | --- | --- | --- |
| Without NLP | 95.13 | 9.753 | 0.9918 | 0.0065 |
| With NLP | 64.83 | 8.051 | 0.9944 | 0.0033 |

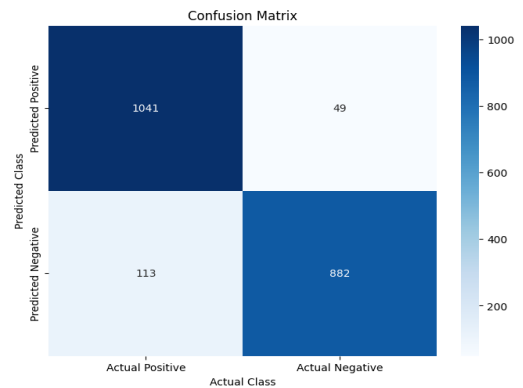

Figure 3: Confusion matrix for sentiment analysis of AMZN sock tweets

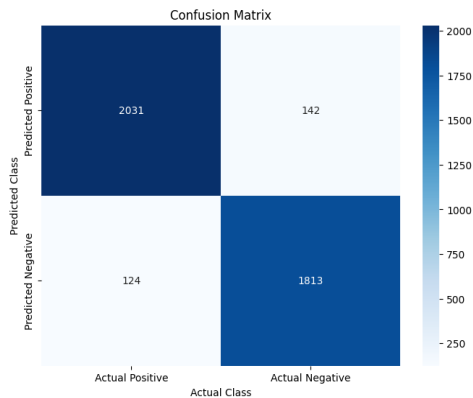

Figure 4: Confusion matrix for sentiment analysis of TSLA sock tweets

The results show that the OHLC + sentimental analysis model outperforms the OHLC-only model in terms of all evaluation metrics for both Amazon and Tesla. The addition of sentimental analysis to the OHLC data improves the prediction accuracy by reducing the MSLE, MSE, and RMSE, and increasing the R-squared value.

## V. CONCLUSION

This research presents an innovative method for stock price prediction of Amazon and Tesla by combining the strengths of CNN and LSTM networks with sentiment analysis of social media data. The results of this study demonstrate that using both OHLC data and sentimental analysis can improve the prediction accuracy of stock prices for Amazon and Tesla. The sentimental analysis provides additional information about market sentiment, which can help to improve the accuracy of predictions. The results suggest that incorporating sentimental analysis into a machine learning model can be a useful approach for improving the accuracy of stock price predictions.

Future work can focus on exploring other machine learning and deep learning models and techniques, such as graph neural networks and attention mechanisms, to further improve the accuracy of stock price predictions.